\documentclass[letterpaper, 10 pt, conference]{ieeeconf}  

\IEEEoverridecommandlockouts                              

\usepackage{graphics} 
\usepackage{graphicx} 
\usepackage{subfigure}
\usepackage{amsmath} 
\usepackage{amssymb}  
\usepackage{algorithm, algorithmic} 
\usepackage{bm}
\usepackage{xcolor}
\usepackage{dblfloatfix}
\usepackage{multirow}
\usepackage{cite} 
\usepackage{array} 
\usepackage{wrapfig} 
\usepackage[table]{colortbl} 

\newcommand{\tflag}[1]{}
\newcommand{\tfflag}[1]{}

\makeatletter
\let\NAT@parse\undefined
\makeatother
\usepackage{hyperref}  

\title{\LARGE \bf
A Small Form Factor Aerial Research Vehicle for Pick-and-Place Tasks with 
Onboard Real-Time Object Detection 
and Visual Odometry
} 

\author{Cora A. Dimmig$^{1}$, 
Anna Goodridge$^{2}$,
Gabriel Baraban$^{1}$, 
Pupei Zhu$^{2}$, 
Joyraj Bhowmick$^{2}$, 
and Marin Kobilarov$^{1}$
\thanks{$^{1}$Department of Mechanical Engineering and the Laboratory for Computational Sensing and Robotics (LCSR), Johns Hopkins University, Baltimore, MD 21218, USA. Email: {\tt\small cdimmig@jhu.edu, marin@jhu.edu}}%
\thanks{$^{2}$LCSR, Johns Hopkins University, Baltimore, MD 21218, USA.}%
}

\begin{document}

\maketitle
\thispagestyle{empty}
\pagestyle{empty}

\begin{abstract}

This paper introduces a novel, small form-factor, aerial vehicle research platform for agile object detection, classification, tracking, and interaction tasks. 
General-purpose hardware components were designed to augment a given aerial vehicle and enable it to perform safe and reliable grasping.
These components include a custom collision tolerant cage and low-cost Gripper Extension Package, which we call GREP, for object grasping. 
Small vehicles enable applications in highly constrained environments, but are often limited by computational resources. 
This work evaluates the challenges of pick-and-place tasks, with entirely onboard computation of object pose and visual odometry based state estimation on a small platform, and demonstrates experiments with enough accuracy to reliably grasp objects. 
In a total of 70 trials across challenging cases such as cluttered environments, obstructed targets, and multiple instances of the same target, we demonstrated successfully grasping the target in 93\% of trials. 
Both the hardware component designs and software framework are released as open-source, since our intention is to enable easy reproduction and application on a wide range of small vehicles. 

\end{abstract}
\section{INTRODUCTION}

In recent years, Uncrewed Aerial Vehicles (UAVs) have become increasingly popular for a wide variety of pick-and-place tasks such as package delivery, agriculture inspection, and warehouse management. Small agile vehicles are particularly advantageous for navigating in highly constrained environments. 
However, a small vehicle limits available sensing and computing options significantly, which makes running complex onboard algorithms such as image processing, motion planning, state estimation, and other intelligent autonomous functionalities challenging. 

\begin{figure}[tbh]
	\centering
	\includegraphics[width=0.68\columnwidth]{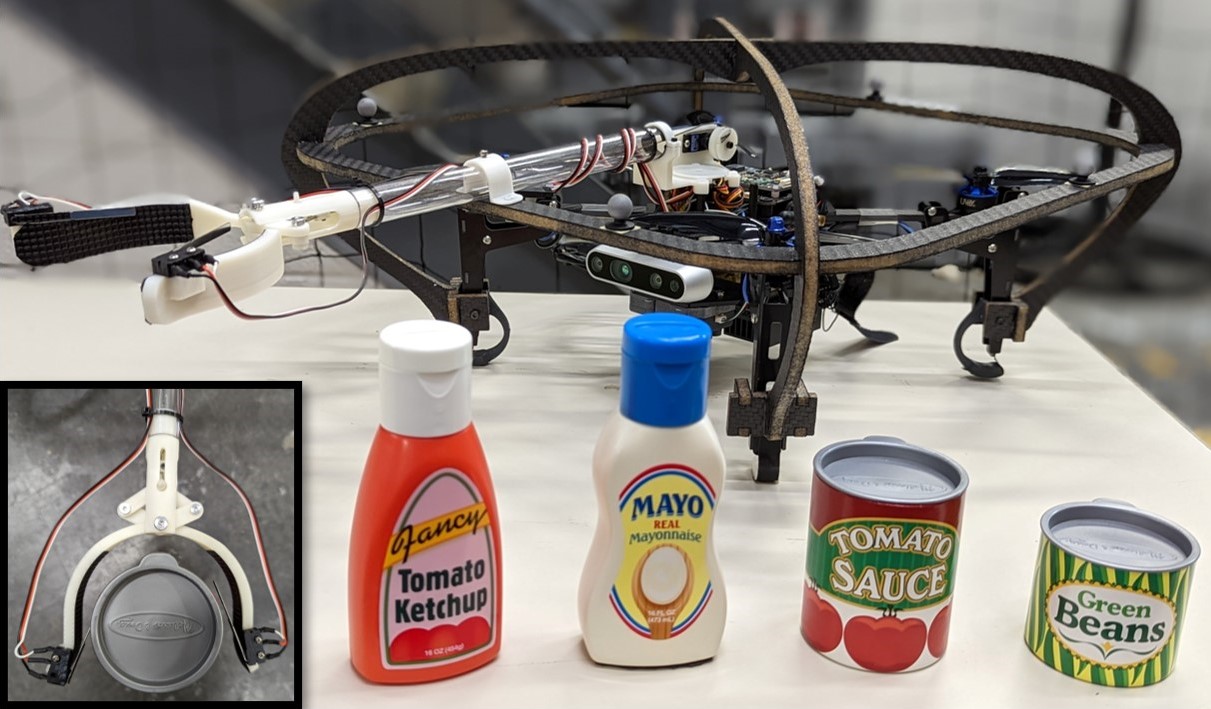}
    \vspace*{-1mm}
	\caption{Aerial research platform, target objects (toy cans and bottles), and view of open gripper, $9.5$~cm across, around $6.5$~cm diameter can.} 
	\label{fig:quad}
	\vspace*{-1mm}
\end{figure}

This paper reports a novel, compact (1.7~kg, 31~cm frame) research UAV for studying aerial grasping applications in constrained environments with entirely onboard computation. 
We modified the UVify IFO-SX quadrotor to be collision tolerant with a carbon fiber foam cage, shock absorbing feet, and a modular Gripper Extension Package, which we call GREP. The fully configured vehicle is seen in Fig.~\ref{fig:quad}. We have open-sourced these designs for use with this vehicle and others. While the reported results are based on the IFO\nobreakdash-SX, the proposed modular designs can be easily modified to enable aerial grasping using other base platforms. 

\begin{figure}[tbh]
	\centering
	\includegraphics[width=0.68\columnwidth]{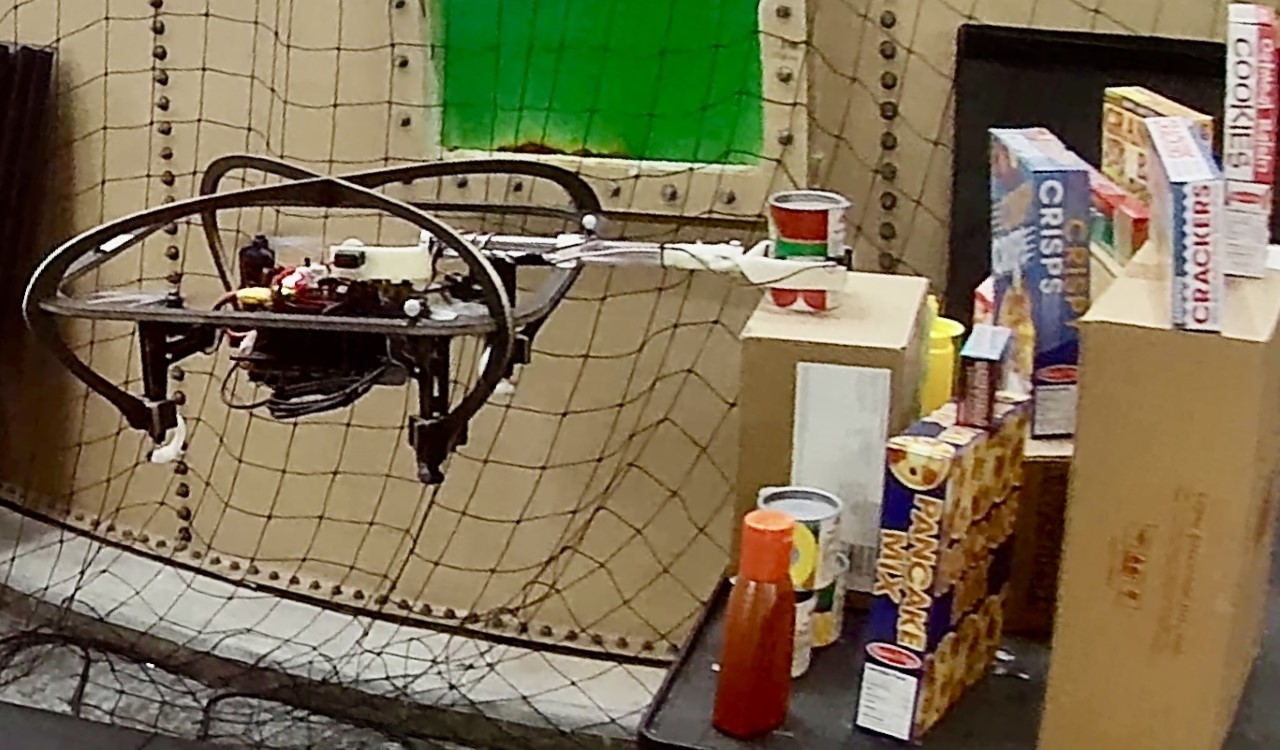}
    \vspace*{-1mm}
	\caption{Autonomous system in flight grasping a can from a cluttered scene.}
	\label{fig:quad_pick}
	\vspace*{-4mm}
\end{figure}

Furthermore, in this work, we present the integration of a task-level fault-tolerant state machine, precision model-based control, visual odometry, and real-time 6D object detection, classification, and tracking. These capabilities are implemented as modular components in an aerial autonomy software framework that can be extended to different tasks. 

We first tested the proposed system in a high fidelity simulation and then in pick-and-place experiments on hardware, as seen in Fig.~\ref{fig:quad_pick}. In an unknown environment, the system detects and tracks an object of interest using onboard visual pose estimation, visually servos to that object and grasps it, then detects a target destination and places the object. 
Our objects of interest were 6.5~cm wide and our gripper opens 9.5~cm wide, which necessitated precise end-effector positioning. 
We evaluated the system's performance in single object environments, cluttered environments, with an obstructed view of the target, and viewing multiple instances of the same object.
A set of 70 pick-and-place hardware experiments in these challenging scenarios yielded 93\% pick action success and 86\% place action success. 

\section{CONTRIBUTIONS}

The objective of this paper is to provide technical details of this newly developed platform, to analyze experimentally its performance, and to discuss challenges towards robust pick-and-place operation in constrained environments. 
This work does not present any novel algorithmic developments. 

\subsection{Vehicle Design Contributions}

Our first objective was to design modular hardware components to enable safe aerial grasping with small UAVs in indoor settings. The contributions include:

\begin{itemize}
    \item Design of a light-weight collision tolerant cage and shock absorbing feet to enhance safety and reliability in constrained environments
    \item Design of a low-cost, lightweight fixed arm with an angular motion gripper, including grip detection circuit
    \item Open-source\footnote{Design Information and Files: \url{https://grabcad.com/library/ifo-sx-cage-and-gripper-extension-package-grep-1}\label{fn:design}} design files for easy reproduction
\end{itemize}

\subsection{System Integration and Evaluation Contributions}

Our second objective was to evaluate performance in pick-and-place experiments. 
Such analysis is useful since few other small form-factor aerial grasping vehicles have been reported 
with entirely onboard computation
for image processing, 6D object pose estimation, state estimation, and motion planning. The contributions include:

\begin{itemize}
    \item New real-time software that integrates established, computationally intensive algorithms on a small platform
    \item High-fidelity simulation for algorithm evaluation and development that can be seamlessly run on hardware
    \item Experimental results demonstrating the robustness of our proposed architecture 
    \item Evaluation of challenges for robust pick-and-place  in constrained environments of increasing levels of complexity in terms of clutter and occlusion
\end{itemize}

\section{RELATED WORK}

\subsection{Aerial Manipulation Platforms}

Aerial manipulation research has been of increasing importance in the last decade~\cite{ollero2021past}. 
Platforms for aerial manipulation are often specialized to the application space \cite{suarez2020compliant:bimanual}, research goals \cite{orsag2017dexterous}, or focused on innovative vehicle or manipulator designs \cite{fishman2021iros:soft, bellicoso2015design, gabrich2018flying}. 
Few standardized, small aerial manipulation platforms are presented due to the variability in requirements for the platform and manipulator.

We sought a generalized research platform to investigate pick-and-place challenges in constrained environments with entirely onboard computation. Our main decision points were: (1) availability of the frame or base platform, (2) size of the vehicle (ideally less than 40~cm frame), (3) payload (for a manipulator and object), and (4) compute capabilities (including a GPU for machine learning). 

In many cases aerial platforms are built around standard frames that have since gone obsolete, such as the Matrice 100 used in \cite{aerialautonomy2021, chermprayong2019integrated, sa2017build} and DJI Flamewheels used in \cite{mohta2018fast, oleynikova2020open}.

Exciting recent work towards a standardized small, agile platform is presented in \cite{foehn2022agilicious}, however this platform is too small to support a manipulator. Conversely, the frames used in \cite{baca2021mrs} are too large for our intended application space.

Ultimately, we determined UVify's IFO-SX met our criteria for a baseline research platform. 
We then 
designed modular, lightweight safety and manipulator additions for this platform that can be easily transferred to other platforms. We offer these designs open-source to the community to aid in entering the aerial grasping space with small vehicles.

Robust flight in constrained spaces requires considering collisions with the environment. Innovative work towards collision tolerant flying robots 
present specialized aerial platforms designed around protective elements, 
as seen in \cite{briod2012airburr, briod2014collision, mulgaonkar2017robust, de2020collision, nguyen2022motion}.
For our reproducible research platform, we designed our cage to fit over an existing frame, such that the design could be adapted for new frames, not be required for flight, and be replaceable if it were to get damaged.

\subsection{Quadrotor Pose Estimation}

Precision control requires robust and accurate pose estimation. 
For this, motion capture systems are widely used, as in \cite{aerialautonomy2021, chermprayong2019integrated, suarez2020compliant:bimanual, gabrich2018flying}. Similarly, GPS is common, as used in \cite{heredia2014control}. However, for real world indoor applications, neither of these approaches would be practical. 
Visual Odometry (VO) methods estimate autonomous vehicles' state using a stream of images captured by onboard cameras \cite{fraundorfer2011visual}. VO approaches have been widely used on aerial platforms, as seen in \cite{foehn2022agilicious, mohta2018fast, baca2021mrs, sa2017build, meier2012pixhawk, oleynikova2020open}. 
In this work, we use a stereo keyframe-based VO approach from the NVIDIA Isaac Software Development Kit (SDK) \cite{isaac20202svio}. 
As part of the NVIDIA Isaac SDK, it is computationally efficient on our onboard Jetson Xavier NX. 

\subsection{Object Pose Estimation}

An integral part of aerial manipulation is the perception of the payload object pose.  Some systems place the object at a known location \cite{appius2022raptor}, or rely on fiducial markers 
for 6-degree-of-freedom (DOF) position tracking relative to an onboard camera, as used in \cite{aerialautonomy2021, heredia2014control, meier2012pixhawk}. 
Motion capture systems can provide sub-centimeter accuracy and are highly prevalent, e.g. in \cite{chermprayong2019integrated}. 
Unfortunately, 
these methods are not applicable to our setting, since we assume that the environment is not instrumented and objects can be at arbitrary locations. 

Visual object pose estimation can be solved through machine learning techniques, e.g. \cite{wohlhart2015learning, hu2019segmentation}, with computational requirements that are typically prohibitive for onboard computation on small UAVs. 
We looked to evaluate 
performance with entirely onboard computation using a common, off-the-shelf algorithm, for which we have not found previous studies on a small platform. For this purpose, we selected Deep Object Pose Estimation (DOPE) \cite{tremblay2018corl:dope}, a neural network for 6-DOF pose estimation of a set of known objects.

\section{HARDWARE}

\subsection{Compact Quadrotor Research Platform}

The IFO-SX has a 31~cm frame, 18~cm propellers, and base weight of 1.38~kg.
In this small form factor it comes equipped with a NVIDIA Jetson Xavier NX for computation, a Realsense D435i for IMU data and color, depth, and stereo infrared images, and a PX4-based autopilot. 
Our weight budget for additions to the platform was 300~g calculated based off of an experimentally evaluated maximum vehicle payload and an additional 100~g for the grasped object. Therefore, weight was a driving 
requirement for all designs. 

We designed and fabricated a custom, carbon fiber foam core cage for safety and robustness to collisions. 
Carbon fiber foam core allows for minimal structure (keeping sensor regions free) and thus minimal weight, while maintaining rigidity. The structure was designed to protect the propellers in case of a crash and in such a way that the cage's 2D contours could be cut and assembled with enough surface area for adhesion with epoxy. 
We developed shock-absorbing feet to strain under the impact load of a hard vehicle landing much like the crumple zone of a car. Capitalizing on a monolithic structure, preexisting mounting points on the vehicle's legs, and the ubiquity of 3D printers, the feet are 3D printed ABS with a final weight of 5g each, deflect up to 5mm before breaking, and are inexpensive enough to be replaced after hard landings.  

\begin{figure}[tbh]
	\vspace*{-7mm}
	\centering
	\includegraphics[width=0.8\columnwidth]{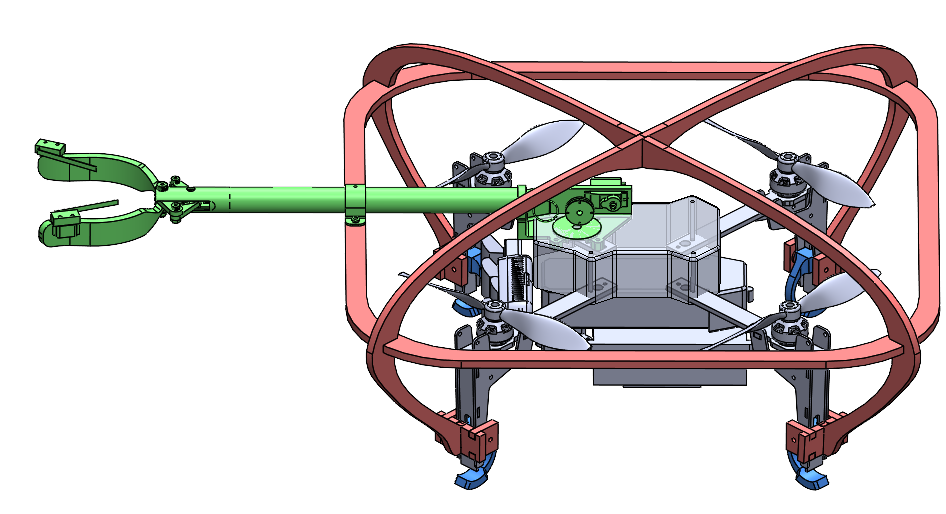}
    \vspace*{-3mm}
    \caption{CAD model of vehicle with integrated modular components.}
	\label{fig:cad}
	\vspace*{-3mm}
\end{figure}

The modified platform fully configured, as seen in 
Fig.~\ref{fig:quad}, has a mass of 1.67~kg. Table \ref{tab:design} and the design files$^{\ref{fn:design}}$ include more detailed information about the custom components. Fig.~\ref{fig:cad} shows the modular components added to the base vehicle in the CAD model. These components were all designed to easily bolt on to an existing platform.

\begin{table}[tbh]
\centering
\vspace*{-1mm}
\caption{Custom Modular Designs for Robustness and Grasping 
}
\label{tab:design}
\vspace*{-3mm}
\begin{center}
\renewcommand{\arraystretch}{1.3}
\begin{tabular}{ >{\centering\arraybackslash}p{3.2cm} | >{\centering\arraybackslash}p{1.3cm} | >{\centering\arraybackslash}p{1.4cm} | >{\centering\arraybackslash}p{1cm} }
\hline
\vspace{0.5mm}\textbf{Component} & \vspace{0.5mm}\textbf{Weight [g]} & \textbf{Dimensions (L$\times$W$\times$H) [mm]} & \vspace{0.5mm}\textbf{Cost [\$]} \\ 
\hline
\hline
& & \\[-3ex]
\includegraphics[scale=0.091]{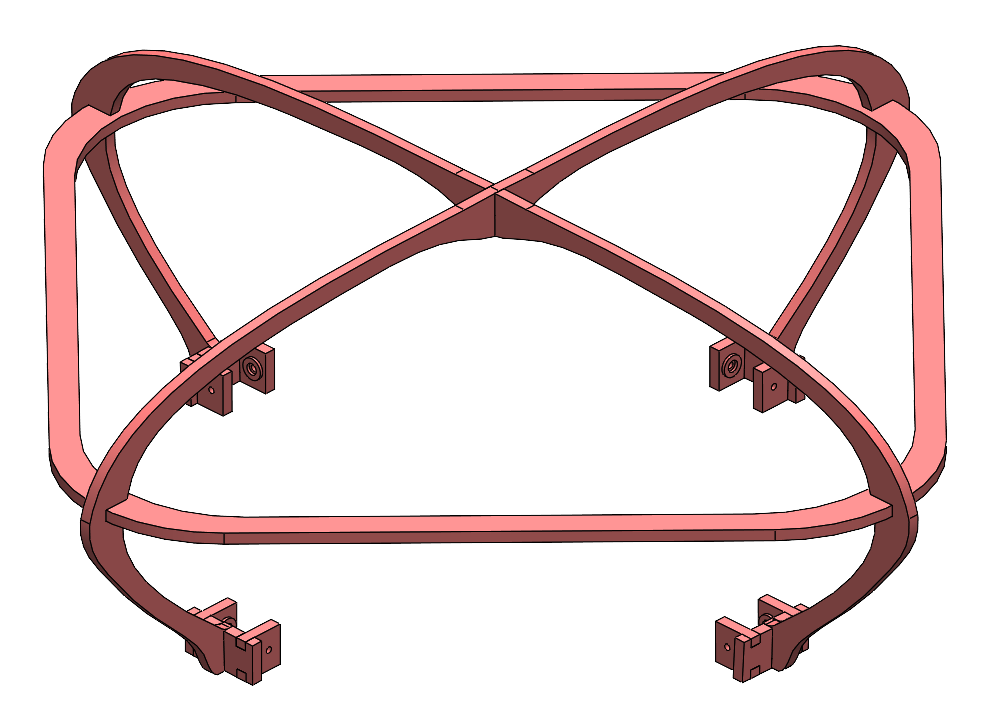} \newline Collision-Tolerant Cage & 162 & 440 $\times$ 440 $\times$ 200 & 516 \\
\hline
& & \\[-3ex]
\includegraphics[scale=0.101]{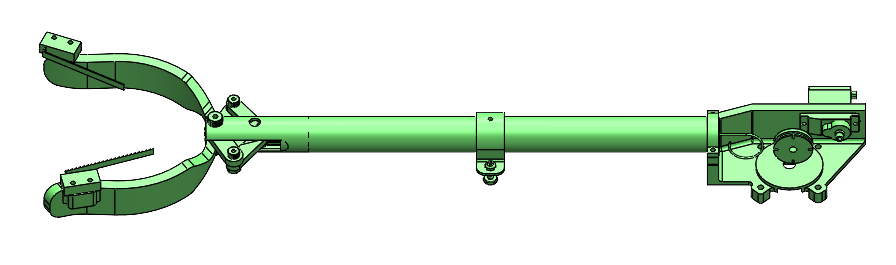} \newline Gripper Extension Package & 91 & 440 $\times$ 120 $\times$ 45 & 43 \\
\hline
& & \\[-2.5ex]
\hspace{4mm}\includegraphics[scale=0.15]{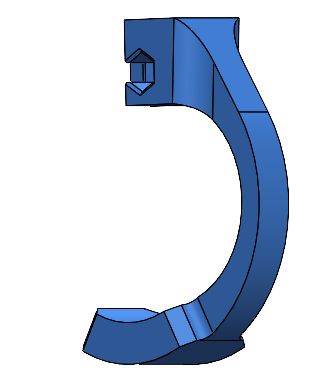} \newline Shock Absorbing Foot & 5 & 36 $\times$ 14 ~~$\times$ 52 & 1 \\
\hline
\end{tabular}
\end{center}
\vspace*{-3mm}
\end{table}

\subsection{Gripper Extension Package}

\begin{figure}[tbh]
    \vspace*{2mm}
	\centering
	\includegraphics[width=0.8\columnwidth]{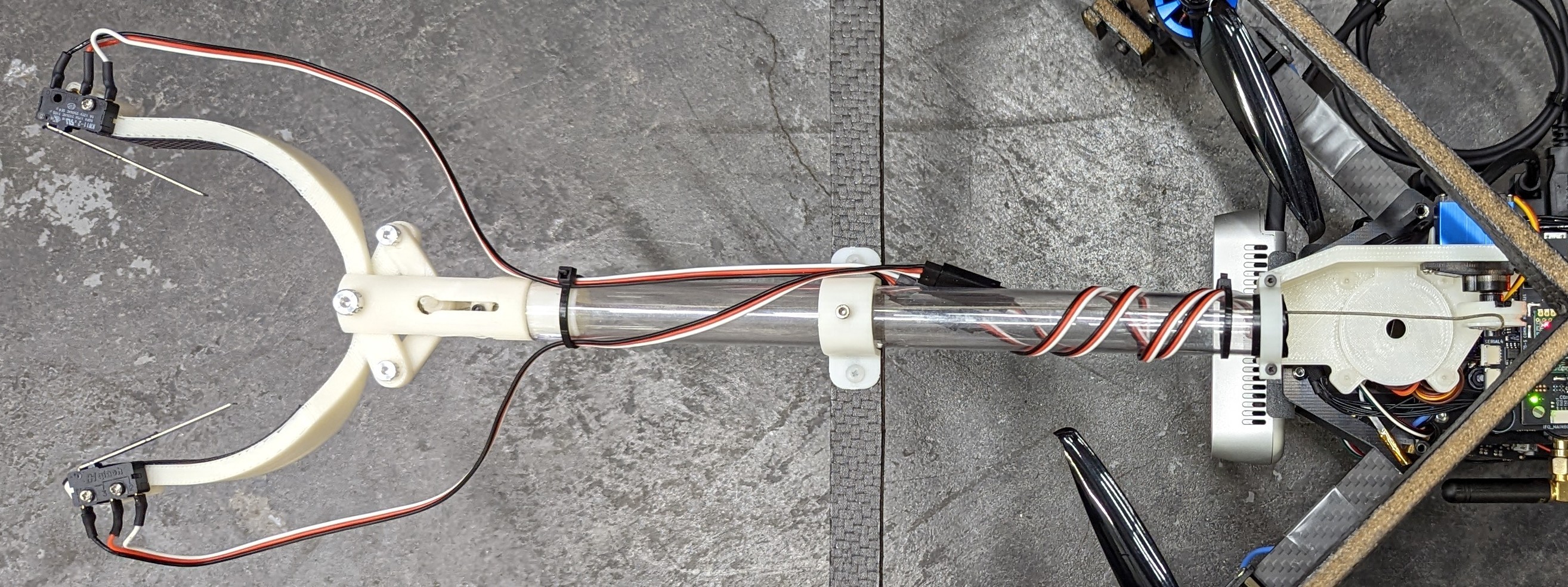}
    \vspace*{-1mm}
    \caption{GREP: Fixed arm with angular motion gripper.}
	\label{fig:gripper}
	\vspace*{-6mm}
\end{figure}

We designed a low-cost, easy to reproduce, lightweight 2-jaw angular motion Gripper Extension Package (GREP) for picking up small objects, as seen in Fig.~\ref{fig:gripper}. 
GREP includes standard hardware and 3D printed ABS mounts, jaws, and linkages. All design files are open-source$^{\ref{fn:design}}$. 
The horizontal design minimizes complexity, enabling a final GREP weight of 91~g, and allows for easy adaptation onto other platforms. 

The length of the extension is set by an acrylic tube that can be swapped depending on the application. On this vehicle we used a 22.5~cm tube for 22~cm of extension from the edge of the cage to the tip of the jaws. 
There is a rubber layer on the inside jaw surface to increase the coefficient of friction between the jaws and the target object.
As mounted in Fig.~\ref{fig:gripper}, the gripper's max payload is 1~kg (11 times GREP's weight) before objects slip out, when tested with a toy can. 

The gripper is actuated by an onboard 
servo motor with a winch mechanism that pulls the gripper closed. 
A spring mounted between the gripper linkages and pivot point assists returning the gripper to its open position.
The servo motor is controlled via a Seeeduino XIAO microcontroller. 
Two snap-action switches at the gripper opening are used to determine if an object is gripped. 
The gripper opens about $9.5$~cm at its widest point and the target objects in this report are about $6.5$~cm in diameter, as seen in Fig.~\ref{fig:quad}. Therefore, high precision control is necessary for successful grasping.

\section{SOFTWARE FRAMEWORK}

\begin{figure}[tbh]
	\vspace*{2mm}
	\centering
	\includegraphics[width=0.85\columnwidth]{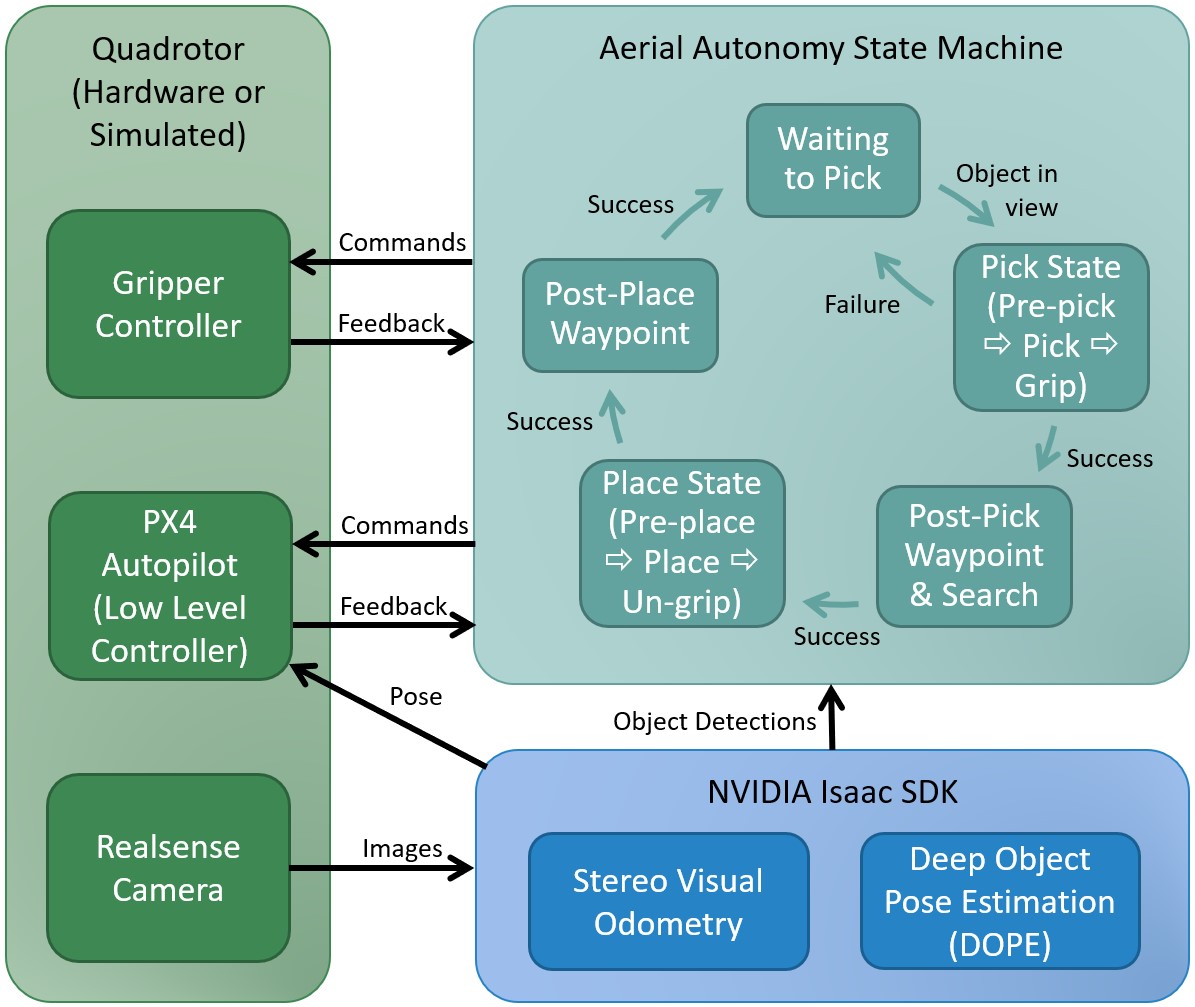}
    \caption{Software flow diagram showing a simplified pick-and-place state machine, begins at ``Waiting to Pick'' state.}
	\label{fig:flow_chart}
	\vspace*{-5mm}
\end{figure}

We designed our software in the Robot Operating System (ROS) 
to be highly modular for integration with different vehicles, manipulators, object detectors, and state estimators. 
Fig.~\ref{fig:flow_chart} shows the key components and information flow. 

\subsection{Aerial Autonomy}

We use the Aerial Autonomy (AA) framework\footnote{\url{https://github.com/jhu-asco/aerial_autonomy}} 
\cite{aerialautonomy2021} 
for task definition. 
We expanded this framework for use with: (1) a PX4-enabled vehicle, (2) a fixed angular motion gripper, (3) vision based object detection, and (4) VO. 
For the evaluation reported herein, we used established object detection and VO algorithms, but  
these high-level contributions to AA allow for easily exchanging hardware and algorithmic components.

Using these additions to the AA framework, we designed a fault-tolerant finite state machine, which continuously checks for controller and hardware failures 
to allow for quick and safe recovery.
For ``high level'' control through Roll-Pitch-Yawrate-Thrust commands, we employ an acceleration-based controller 
to track a polynomial reference trajectory as defined in \cite{aerialautonomy2021}. 
The 9th degree polynomial reference trajectory starts from the vehicle's position and yaw at initialization and terminates at a final position and yaw determined by the task, such as relative to an object of interest. 

\subsection{Aerial Autonomy State Machine}\label{section:aa_state_machine}

Fig.~\ref{fig:flow_chart} depicts a simplified version of the AA state machine. The vehicle starts with the object of interest in the field of view of the camera. AA begins tracking the object based on the detections from DOPE. 
Mirroring the object's yaw, 
the vehicle flies to relative pre-pick and pick positions, resulting in the gripper encircling the target object and commands the gripper to close. When approaching the pick position the gripper will begin to obstruct detection of the target object and the last filtered estimate of the object's location is used. 

The system is continuously checking for system faults, such as the target object not being grasped, tracking of the object timing out, and error in the vertical axis or rotational error being above specified thresholds when entering the pick grasping region. If a fault is encountered, the vehicle resets to its original pose and looks for the object of interest again. 

If the target object is detected as successfully grasped, the vehicle moves to a post-pick waypoint, with respect to the pick position, and begins looking for the destination object. We implemented a simple search routine for the vehicle to incrementally move downwards if the destination object is not detected. Once the destination object is found, the vehicle goes to relative pre-place and place positions and releases the target object to the right of the destination object. 

\subsection{Quadrotor Pose Estimation}

We use stereo VO from NVIDIA's Isaac SDK \cite{isaac20202svio} for pose estimation of the vehicle. 
We selected this approach since it is an established algorithm optimized for our onboard computer. The VO approach operates on the stereo infrared images from the Realsense D435i, with the emitter turned off, and uses the Elburs VO library for calculation of the 3D pose of the vehicle. Running onboard the Jetson Xavier NX with the rest of our software stack, the VO publishes at about 15~Hz. We experimentally determined this to be insufficient for precision control and calculating velocities.

To improve the state estimation, we fuse the VO estimate with the onboard PX4 Extended Kalman Filter (EKF), which we publish more reliably at about 40~Hz. 
We are operating indoors without GPS, so the fused PX4 state estimation comes from the onboard VO, gyroscope, accelerometer, and magnetometer. 
The high level controller uses this fused state estimate to calculate Roll-Pitch-Yawrate-Thrust commands, which are then passed to the low level PX4 controller.

\subsection{Object Pose Estimation and Grasp Strategy} 
\label{sec:object_detect_grasp}

For 6-degree-of-freedom (DOF) pose estimation and classification of objects, we use DOPE \cite{tremblay2018corl:dope}. DOPE predicts the pose of known classes of objects from a single RGB image. We have integrated our code base with the instantiation of DOPE in NVIDIA's Isaac SDK \cite{isaac20202dope} for an off-the-shelf baseline of entirely onboard computation. 
We pass Realsense camera imagery and object detection and classification information through a custom Isaac ROS bridge.

We use the object detections from DOPE for object tracking in AA. 
To refine the poses provided by DOPE, we remove invalid poses and filter the poses. 
We remove invalid poses by checking if the pose is in the camera's field of view, or frustum. We do this by projecting the pose into pixel coordinates, using the camera intrinsic matrix, and evaluating if that pixel is within the bounds of the image. For focal length $(f_x, f_y)$ and principal point $(c_x, c_y)$, the point $(x,y,z)$ can be projected to $(u, v)$ as follows.
\begin{align}
    (u, v) &= \frac{1}{z} \bigg(f_x x + c_x z, ~f_y y + c_y z\bigg)
\end{align}
Then we evaluate if $(u,v)$ is within the bounds of the image, 
if not we consider this detection invalid.

Each valid detection is evaluated against the current filtered object estimates, by thresholding distance, to determine if the new detection is a new object or an updated detection of a known object.
If it is grouped as a detection of a known object, we add it to the filter for that object.
Since DOPE does not provide covariances for Kalman Filtering, we exponentially filter the position and yaw of each detected object with a decaying gain. 
The gain decays over $T$ steps and then remains fixed at the desired gain $\gamma_d$. At step $t$, when $t < T$, the current gain, $\gamma_t$, is $\gamma_t = 1 - (1 - \gamma_d) \frac{t}{T}$.
For $t \geq T$, $\gamma_t = \gamma_d$. 
Then the filtered estimate $p_t$, comes from the previous filter estimate $p_{t-1}$, and the current measurement from DOPE $m_{t}$, as follows.
\begin{align}
    p_t = p_{t-1} + \gamma_t (m_{t} - p_{t-1})
\end{align}

The closest target object to the vehicle is then identified and we employ a visual servoing controller using the object's filtered pose to maneuver the UAV to target poses relative to the object. 
As a basic grasp strategy, these target poses, which are tuned to optimize performance, are defined as relative transforms from the object that result in the gripper surrounding the object of interest, i.e. the vehicle maneuvers to a position offset from the object by the transform from the vehicle center to the gripper center, such that when the vehicle reaches that target pose, the object will be in the center of the gripper. When the error to the desired pose is within user-defined thresholds, the gripper closes.

As representative small objects, we use the Household Objects for Pose Estimation (HOPE) models with DOPE 
\cite{tyree2022:hope}. 
Fig.~\ref{fig:quad} shows the four objects we used in this evaluation. The two target objects, Mayo and Tomato Sauce, and the corresponding destination objects, Ketchup and Green Beans. Fig.~\ref{fig:scenarios} shows onboard detections of the target objects. 
We found detecting two objects at a time to be the processing limit on our Jetson Xavier NX, with GPU usage up to 99\%. 

\subsection{High Fidelity Simulation}

\begin{wrapfigure}{o}{0.42\columnwidth}
    \vspace*{-4mm}
    \includegraphics[width=0.41\columnwidth]{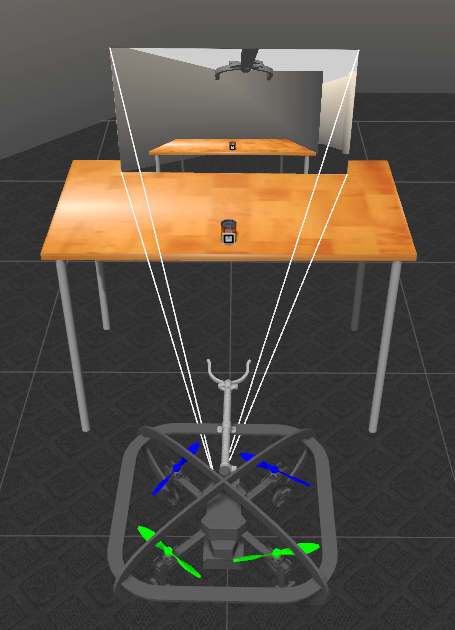}
    \caption{Gazebo simulation.}
    \label{fig:quad_can_sim}
    \vspace*{-3mm}
\end{wrapfigure}

We built a simulation environment in Gazebo for rapid testing and prototyping of algorithms. This simulation uses PX4's Software-In-The-Loop (SITL) simulation architecture
and allows for software to be easily tested on a simulated vehicle that can be directly swapped for a real vehicle. 
We tuned the controller gains in simulation and found only minor adjustments were required on the real vehicle. 

Fig.~\ref{fig:quad_can_sim} shows the quadrotor in the Gazebo simulation environment in front of a target object on the table. The simulated camera's view is shown at the end of the camera's frustum. AR tags are used in simulation for object detection. The simulated environment was essential for the deployment and tuning of the onboard visual servoing control strategy.

\section{PICK-AND-PLACE EXPERIMENTS}

\subsection{Experiment Setup}

\begin{figure*}[tbh]
	\vspace*{2mm}
	\centering
	\includegraphics[width=0.59\paperwidth]{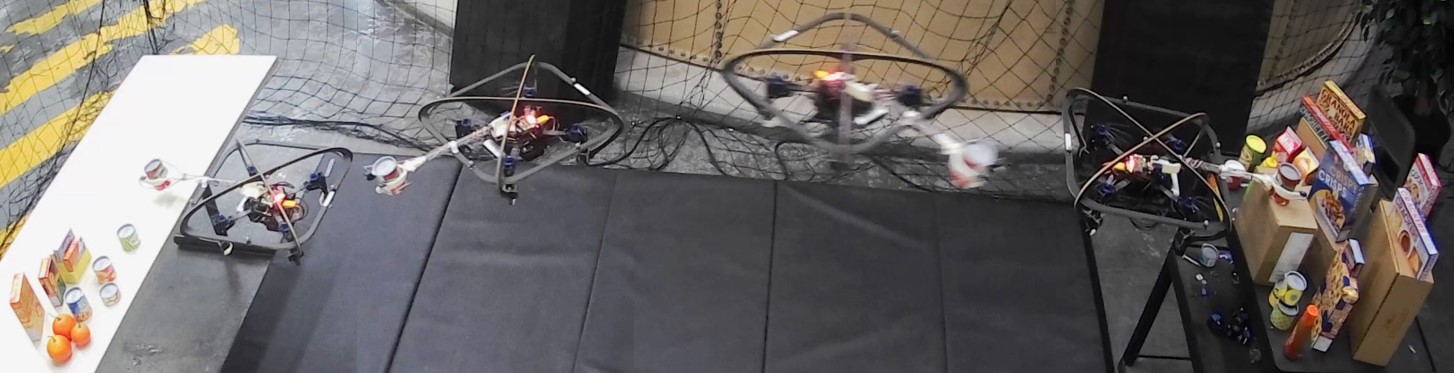}
    \vspace*{-2mm}
	\caption{Aerial time lapse of a cluttered environment experiment. Pick location on the right and place location on the left.}
	\label{fig:aerial_pick_place}
	\vspace*{-2mm}
\end{figure*}

We performed hardware pick-and-place experiments to evaluate our system's performance. We placed a target object on a cart and a destination object on a table about 3 meters away, as seen in Fig.~\ref{fig:aerial_pick_place}. We manually controlled the system to a variable starting location facing the target object and then switched into autonomous mode, which follows the AA state machine, as described in Section \ref{section:aa_state_machine}.
 
See included video for examples of the experiments\footnote{Pick-and-place experiments: \url{https://youtu.be/XAHcYrbYhy0}}.

\subsection{Performance Evaluation}

\begin{figure*}
    \centering
    \subfigure[Single Bottle]
        {\includegraphics[width=0.175\textwidth]{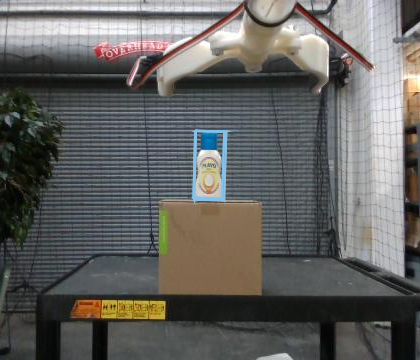}} 
    \subfigure[Single Can]
        {\includegraphics[width=0.175\textwidth]{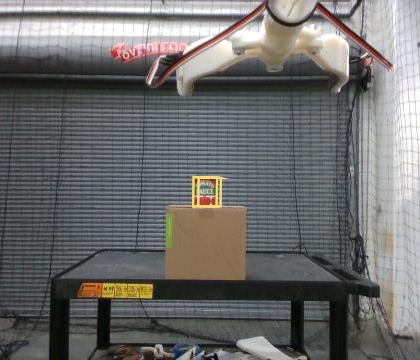}} 
    \subfigure[Can Clutter]
        {\includegraphics[width=0.175\textwidth]{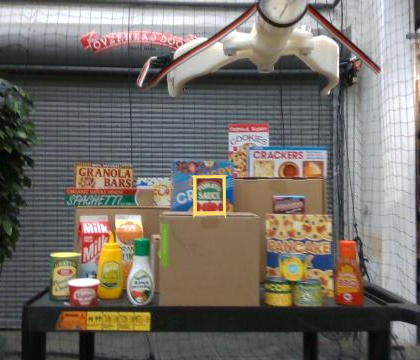}}
    \subfigure[Obstructed Can]
        {\includegraphics[width=0.175\textwidth]{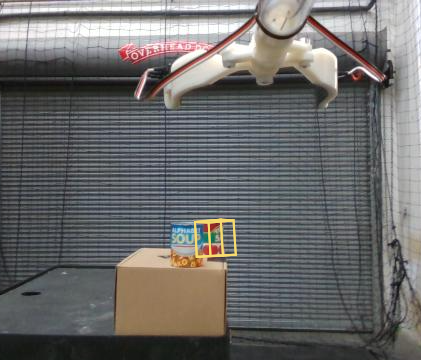}}
    \subfigure[Multiple Can Instances]
        {\includegraphics[width=0.175\textwidth]{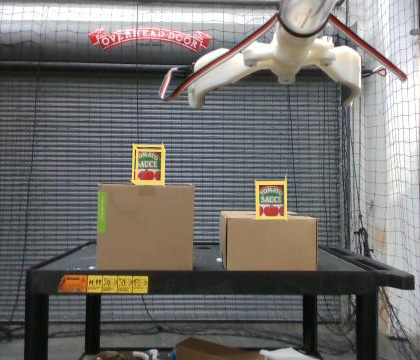}}
        \vspace*{-2mm}
    \caption{Experiment scenarios as viewed from the vehicle's onboard camera, including detected bounding box of target objects.}
    \label{fig:scenarios}
    \vspace*{-5mm}
\end{figure*}

To evaluate the system's performance, we ran 70 pick-and-place experiments from variable initial conditions. 
The pick action was considered a success if the target was gripped. 
The place action was considered a success if the target was released on the table to the right of the destination object. 

In the 70 pick-and-place experiments, we tested the following scenarios: single object, clutter, obstruction, and multiple instance, as depicted in Fig.~\ref{fig:scenarios}. In the single object trials, the object of interest was in a relatively uncluttered environment. We ran 20 trials each of two classes of objects: cans and bottles. 
As a challenge scenario, we performed an additional 10 trials in a cluttered environment with many different items, some visually similar, surrounding the target object and destination object. We ran another 10 trials with an object obstructing 10-30\% of the target when viewed from the vehicle's initial position. The target was rotated which required the vehicle to shift laterally to approach at an angle and avoid the obstruction. 
Lastly, we ran 5 trials with multiple instances of the target object. The system needed to identify two instances of the object, filter separate pose estimates, and then pick-and-place each successively; we count these trials as two pick-and-place experiments. 
An additional offset is added to the place destination for each instance to assure they are not placed co-located. 
Cans were used as the target object for the challenging scenarios of clutter, obstruction, and multiple instances.
Clutter, obstructions, and multiple instances of objects are common points of failure for VO and object detection approaches, so we wanted to evaluate baseline performance with common, off-the-shelf algorithms running entirely onboard. 

\subsection{Performance Metrics}

Table \ref{tab:metrics} shows performance metrics from each category of experiments. Our overall pick success rate across all 70 experiments was 93\% and the place success rate was 86\%. The number of trials the system needed to reset at least once during, due to system faults during the pick action, are tracked in Table \ref{tab:metrics}. The system never reset during the place action. We considered a success rate of 90\% or higher to be ideal performance and are shown in green, 85-90\% is shown in yellow, and below 85\% is in red.

\begin{figure}
    \vspace*{2mm}
    \centering
    \subfigure
        {\includegraphics[width=0.78\columnwidth, trim = 0cm 0cm 0cm 0.7cm, clip]{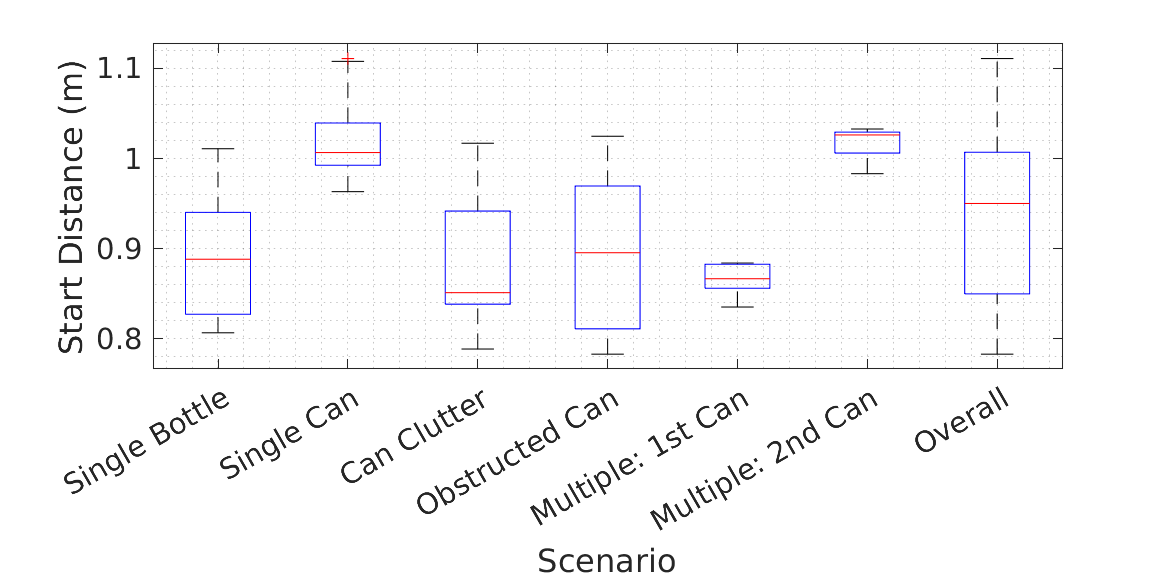}}
    \subfigure
        {\includegraphics[width=0.78\columnwidth, trim = 0cm 0cm 0cm 0.7cm, clip]{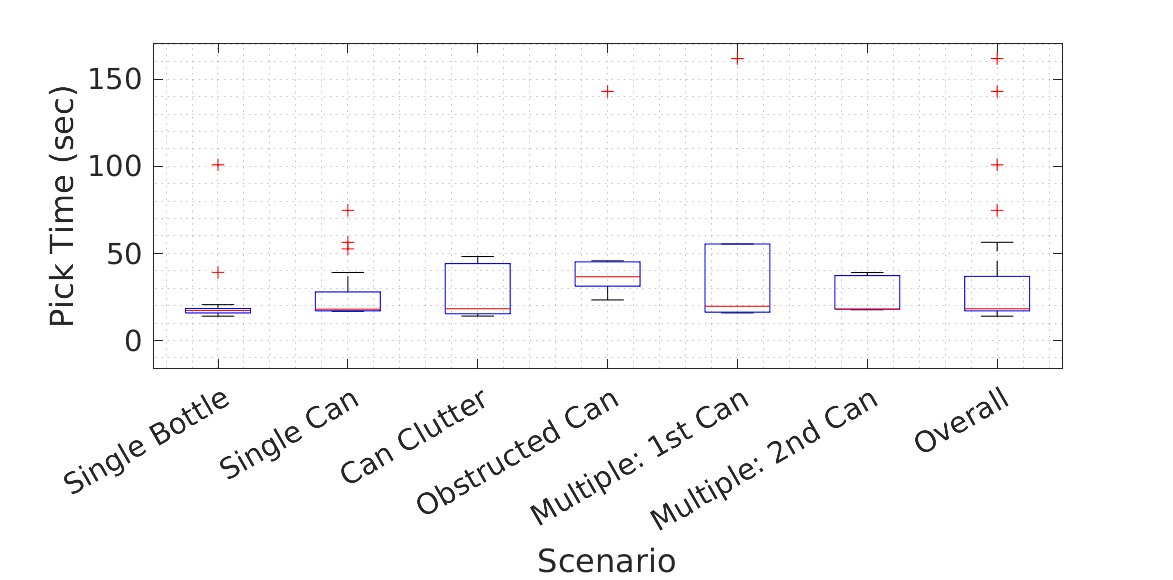}} 
    \vspace*{-2mm}
    \caption{Metrics for starting distance (top) and total time for the pick action (bottom) for each experimental scenario.}
    \label{fig:start_dist_pick_time}
    \vspace*{-4mm}
\end{figure}

\begin{table}[b]
\vspace*{-4mm}
\caption{Pick-and-Place Performance Metrics}
\label{tab:metrics}
\vspace*{-3mm}
\begin{center}
\renewcommand{\arraystretch}{1.3}
\begin{tabular}{ c | c c | c }
\hline
\multirow{2}{*}{\textbf{Scenario}} & \multicolumn{2}{c|}{\textbf{Success Rates}} & \multicolumn{1}{c}{\textbf{Pick Resets}} \\
& \textit{Pick} & \textit{Place} & \textit{Num Trials} \\ 
\hline
\hline
Single Bottle & \cellcolor[HTML]{abebc6}18/20 & \cellcolor[HTML]{f9e79f}16/18 & 3/20 \\
Single Can &  \cellcolor[HTML]{abebc6}20/20 & \cellcolor[HTML]{abebc6}19/20 & 5/20 \\
Can Cluttered Environment &  \cellcolor[HTML]{abebc6}9/10 & \cellcolor[HTML]{f9e79f}8/9 & 3/10\\
Obstructed Can & \cellcolor[HTML]{f5b7b1}8/10 & \cellcolor[HTML]{f5b7b1}4/8 & 3/10 \\
Multiple Instance: First Can & \cellcolor[HTML]{abebc6}5/5 & \cellcolor[HTML]{abebc6}5/5 & 1/5 \\
Multiple Instance: Second Can & \cellcolor[HTML]{abebc6}5/5 & \cellcolor[HTML]{f5b7b1}4/5 & 2/5 \\
\hline
Overall & \cellcolor[HTML]{abebc6}65/70 & \cellcolor[HTML]{f9e79f}56/65 & 17/70\\
\hline
\end{tabular}
\end{center}
%
%
\vspace*{-1mm}
\caption{Pick Action Overall Metrics}
\label{tab:overall_metrics}
\vspace*{-3mm}
\begin{center}
\renewcommand{\arraystretch}{1.3}
\begin{tabular}{ c c | c }
\hline
\multicolumn{2}{c|}{\textbf{Metric}} & \textbf{Value} \\
\hline
\hline
\multirow{3}{9em}{Starting Distance (m)} & Min & 0.78 \\
& Mean & 0.93 \\
& Max & 1.11 \\
\hline
\multirow{4}{7em}{Pick Time (sec)} & Min & 13.9 \\
& Median & 18.2 \\
& Mean & 29.6 \\
& Max & 161.8 \\
\hline
\end{tabular}
\end{center}
\vspace*{-3mm}
\end{table}
Table \ref{tab:overall_metrics} shows metrics for the pick action across all 70 trials. This includes the starting distance (defined from the vehicle's initial pose to the final filtered estimate of the target object) and the time to perform the pick action (moving from the initial pose to grasping the target object). 
These metrics 
are further broken down by experiment scenarios in Fig.~\ref{fig:start_dist_pick_time}. In each boxplot the center line is the median, the box spans from the 25th to 75th percentiles and the whiskers show the bounds, excluding the outliers denoted by plus signs (+). The intention was to test the system for relatively similar starting distances across each scenario. Additionally, the pick time plot shows that the time to perform the pick action did not vary significantly across the more challenging scenarios.  

\begin{figure}[tbh]
	\centering
    \includegraphics[width=0.735\columnwidth, trim = 1.4cm 4.7cm 2.2cm 5cm, clip]{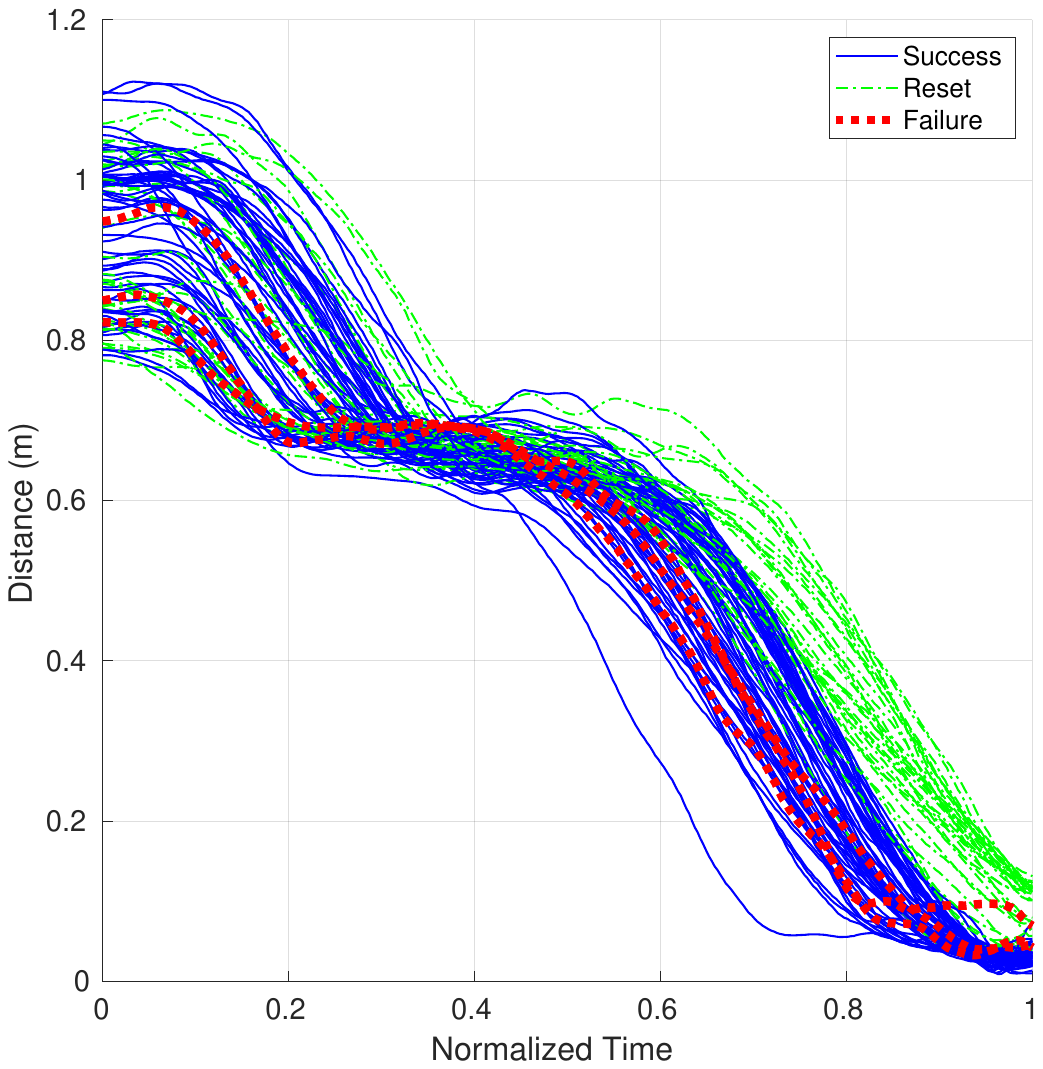}
    \vspace*{-2mm}
	\caption{Distance from gripper center to target object during pick action versus normalized time across 60 trials with direct approaches (excludes obstruction trials).}
	\label{fig:dist_time}
	\vspace*{-6mm}
\end{figure}

Fig.~\ref{fig:dist_time} shows the distance in meters from the gripper center to the final estimate of the target object's pose during the pick action on the Y~axis versus normalized time on the X~axis. This figure includes the 60 trials where the vehicle initializes with the target object along the axis of the vehicle's arm. The obstruction trials were excluded in this figure since the vehicle needed to translate laterally before approaching the target object. In Fig.~\ref{fig:dist_time}, each blue solid line represents a successful pick action, red dotted lines represent a failed pick action, and green dash-dot lines represent each approach to the target object that resulted in an autonomous reset due to a system fault. 
After an autonomous reset the system would reattempt the pick action. These additional attempts are included in Fig.~\ref{fig:dist_time} as new approaches and are tracked with the appropriate coloration and line type.
The points on the Y~axis represent the various starting distances from the object. In each trajectory there is a  plateau around 0.65m to the target object for the pre-pick position, before continuing on to the pick position. 
In  Fig.~\ref{fig:dist_time}, it is desired for the distance to converge to zero, indicating the center of the gripper and the target object becoming coincident. The gripper closes when the pose of the vehicle is within specified tolerances of the target object. In these experiments, those tolerances were 3~cm laterally, 2~cm along the axis of the arm, and 2~cm vertically.

The system fault detection imposes bounds on a grasp region. This can be seen in Fig.~\ref{fig:dist_time} by the successful grasps clustering around a lower terminal distance than the cluster of resets. 
The three pick failures in this plot fall in the region overlapping the successful approaches and reset approaches. 

\section{DISCUSSION}

\subsection{Failure Modes}

\begin{figure}[t]
	\vspace*{1.5mm}
	\centering
    \includegraphics[width=0.95\columnwidth, trim = 0cm 0cm 0cm 1.0cm, clip]{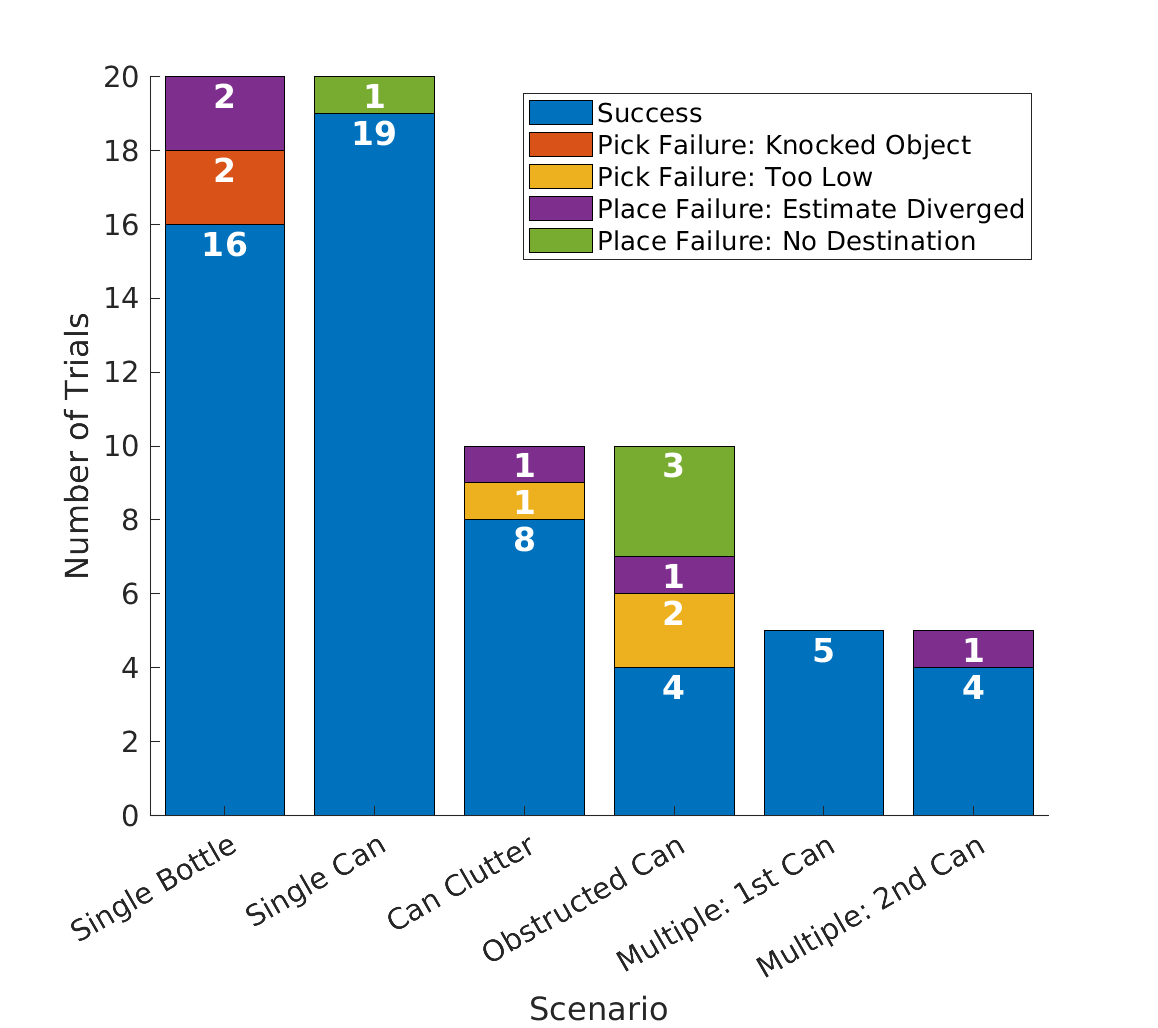}
    \vspace{-2mm}
	\caption{Failure modes across each experimental scenario.}
	\label{fig:failure_modes}
	\vspace*{-4mm}
\end{figure}

Fig.~\ref{fig:failure_modes} depicts the main failure modes encountered in the pick-and-place experiments. There is a bar for each scenario type. The total length of the bar is the total number of experiments in that category. The blue regions of each bar are the trials with successful pick and place actions. Each bar is further divided by the types of pick failures and place failures encountered, which are detailed in the following sections.  

\subsubsection{Pick Action Failures}

We encountered two main failure modes in the pick action: knocking the target object over, which occurred twice, and approaching the target object from too low, which occurred three times. Both instances of knocking the target object over occurred during the single object bottle trials. In one trial the object was gripped too high, which caused it to fall. In another trial the object was bumped during approach. The bottles are less stable than the cans making them more prone to this type of failure. 

Pick action failures due to approaching the object from too low occurred in two obstruction trials and one clutter trial and resulted in the gripper colliding with the box the target object was standing on. In cases, such as these, where the vehicle collided with the environment, the protective cage and shock absorbing feet prevented damage to the vehicle.  

Both of these pick failure modes could be due to the quality of the detection of the object or the specification of the pick grasp region being too large. Tighter constraints on the grasp region could potentially avoid some of these issues. Additionally, greater spatial awareness of objects in the environment could mitigate these issues. The current system acts solely on the tracking information of the object. 

\subsubsection{Place Action Failures}

The two types of place action failures encountered were due to the vehicle state estimation diverging and the destination object not being found. In cases where the destination object was not found the vehicle was too low to see the object in 3 cases and too far back in 1 case, due to human error setting up the environment. This could be mitigated by an improved search routine. Currently the vehicle searches downwards if the destination object was not found. This occurred in 17 trials, including all 4 place failures where the destination object was not found. In all other trials the search routine resulted in the destination object being found. This search routine could be expanded to more thoroughly search the environment. In this work, we looked to demonstrate a pick-and-place capability, leaving more thorough search behaviors to future work. The obstruction trials accounted for 3 of the 4 failures due to not detecting the destination object. This was likely due to the major differences in the testing setup. In the obstruction trials, due to the setup of the environment, the vehicle's position after the pick action was shifted further from the place location in comparison to all other trials.

The other type of place action failure was due to the state estimation diverging. Two components contributed to this behavior: the visual odometry drifting and the vehicle's magnetometer failing. The magnetometer encountered many system faults resulting in failed innovation consistency checks, making the data highly unreliable. The vehicle was only launched when the magnetometer passed innovation consistency checks, but over time could accumulate more error resulting in issues during the place action procedure. Replacing, disabling, or having redundant magnetometers could all improve performance. Furthermore, the visual odometry would occasionally drift and diverge from the fused EKF state measurement. This occurred most often after any sharp movements, such as when the vehicle turned quickly from the pick location to the place location, with an object in the gripper partially occluding the camera's field of view. 
More fault monitoring could identify these issues and potentially allow for autonomous recovery. 

\subsubsection{General Failure Modes}

The quality of the object detection significantly impacts the system's performance. Many erroneous detections of the target objects were filtered out or rejected, as described in Section \ref{sec:object_detect_grasp}. 
Qualitatively, the cans were found to detect more consistently and frequently than the bottles. Occasionally, truly false detects would occur such as detecting a Ketchup bottle on the logo of the Mayo when the Mayo was in the gripper, potentially due to the red colors in the Mayo logo at that proximity and angle. While it was not encountered in the trials reported herein, this phenomenon could have resulted in placing the object in an incorrect location. 

\subsection{Robust System Performance}

Overall, the system performed reliably across a large number of trials of two different classes of objects and challenge cases including a cluttered environment, an obstructed target, and multiple instances of the same object, which required disambiguation of the instances and maintaining operation through two pick-and-place trials. The pick task is the most challenging, requiring precise control of the vehicle, and only encountered 5 failures across 70 trials with relatively comparable performance in the single object cases and smaller challenge scenarios. Performance could be further improved by refining the pick grasp region and incorporating more spatial awareness of the environment. In the obstruction case, the system was able to identify the object and approach from a different angle determined by the rotation of the target object. With greater spatial awareness, the correct approach direction could be determined regardless of the target object's setup orientation. 

Through sensor improvement, further fault mitigation, and improved search behaviors, the place action performance could be further improved as well. Furthermore, to improve overall performance and reliability, the system's gains could be further tuned to improve trajectory tracking accuracy.

\section{CONCLUSION}

In this paper we present a novel small UAV for aerial grasping research with entirely onboard sensors and computation. We provide technical details for a collision-tolerant cage, shock-absorbing feet, and a lightweight, low-cost Gripper Extension Package (GREP). We demonstrate pick-and-place experiments successfully running DOPE and VO entirely onboard and using object pose estimates for visual servoing with enough accuracy that small objects were able to be grasped with a 93\% success rate and placed at a target destination with a 86\% success rate in a wide variety of challenging scenarios. Moving forward we look to more tightly couple the object pose estimation with perception information and refine the control strategy for these grasping tasks in order to further improve accuracy and agility 
and increase the general applicability of this work. 



\section*{ACKNOWLEDGMENT}
We gratefully acknowledge the support of
the National Science Foundation under grant \#1925189.
We are grateful to Austin McWhirter who contributed to the hardware experimentation. 
\bibliographystyle{IEEEtran}
\bibliography{references}

\end{document}